%% file: main.tex
\documentclass[10pt,twocolumn,letterpaper]{article}

\usepackage[pagenumbers]{wacv} 


\usepackage[accsupp]{axessibility}  
\definecolor{wacvblue}{rgb}{0.21,0.49,0.74}
\usepackage[pagebackref,breaklinks,colorlinks,allcolors=wacvblue]{hyperref}

\usepackage[table]{xcolor}

\def\wacvPaperID{281} 
\def\confName{WACV}
\def\confYear{2027}

\title{Painting Change Map as guidance for generating painting processes}

\author{
    Markus Pobitzer \quad
    Chang Liu \quad
    Niccolò Biondi \quad
    Nicu Sebe \\[0.3cm]
    University of Trento, Italy \\
    {\tt\small markus.pobitzer@unitn.it}
}

\begin{document}


\twocolumn[{%
\renewcommand\twocolumn[1][]{#1}%
\maketitle
\begin{center}
    \centering
    \captionsetup{type=figure}
    \includegraphics[width=.99\textwidth]{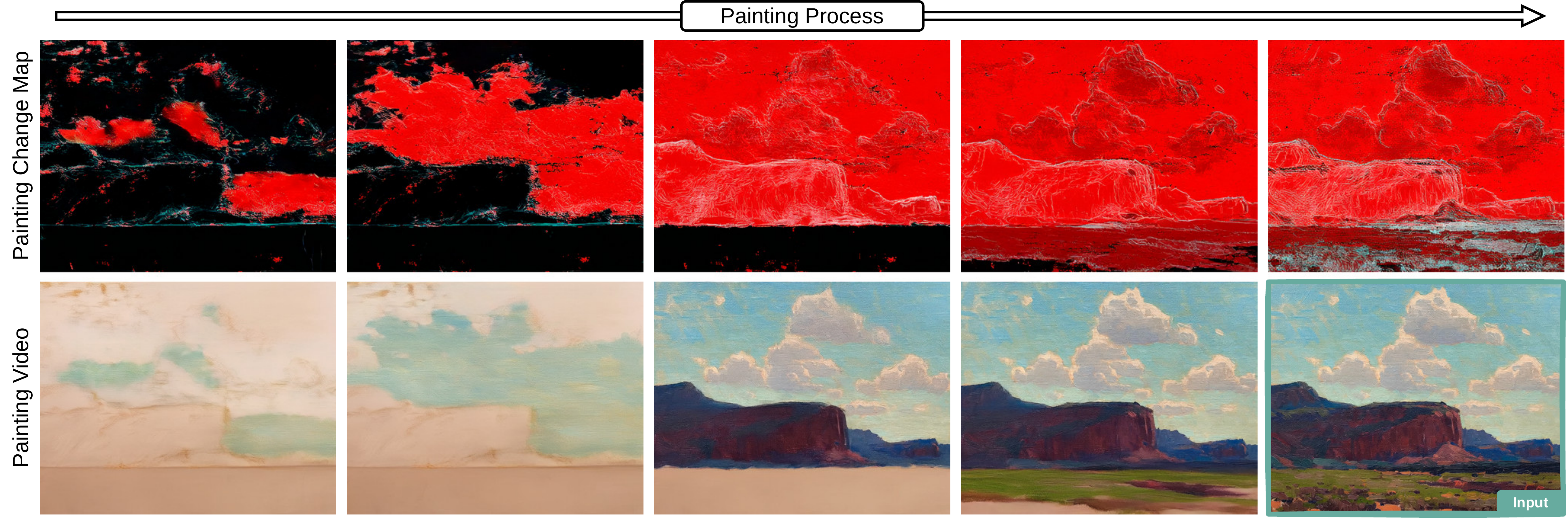}
    \captionof{figure}{Painting Change Map (PCM). We propose a novel two-step pipeline for general painting video reconstruction. Given a final input image (far right), our method first predicts a sequence of changes applied to the canvas. This map then serves as temporal and spatial guidance to synthesize a realistic video of the painting process. All intermediate frames shown are entirely generated by our model. \textit{Continental Divide} by Edgar Alwin Payne, Public Domain, via Wikimedia Commons.}
    \label{fig:teaser}
\end{center}%
}]

\input{sec/0_abstract}    
\input{sec/1_intro}
\input{sec/2_related_work}
\input{sec/3_method}

\input{sec/4_experiments}
\input{sec/5_conclusion}
\clearpage

{
    \small
    \bibliographystyle{ieeenat_fullname}
    \bibliography{main}
}

\input{sec/6_supplementary}

\end{document}

%% file: sec/0_abstract.tex
\begin{abstract}
Digital art generation has undergone a massive paradigm shift, moving from static image synthesis to dynamic video generation that can mimic the human painting process. 
While existing models can synthesize visually stunning time-lapses of paintings being created or "unpainted," they largely treat the creative process as a black box. Current conditioning mechanisms are either too rigid, relying on static geometric cues like Canny edges or depth maps, or too holistic, utilizing simple scalar progress variables that offer no localized control. Consequently, users lack the agency to steer how a painting evolves over time. In this work, we argue that true control over the video generation of artwork requires a joint understanding of two fundamental dimensions: where an artist acts, and when those actions take place. To achieve this, we introduce the Painting Change Map, a novel spatio-temporal control mechanism that explicitly tracks normalized structural changes across a 1,000-timestep continuum, a resolution sufficient to capture fine-grained stroke-level changes while remaining computationally tractable. By mapping localized visual differences into an encoded RGB space, our representation enables fine-grained guidance for both the forward painting and reverse unpainting processes. Extensive benchmarking demonstrates the efficacy of our approach. We outperform existing control baselines and achieve competitive or superior results compared to state-of-the-art painting methods. Furthermore, under oracle conditions, our method surpasses all baselines by a significant margin. Code is available at \url{https://github.com/Markus-Pobitzer/painting-change-map}.


\end{abstract}

%% file: sec/1_intro.tex
\section{Introduction}
\label{sec:intro}

The creation of art is as much about the temporal process as it is about the final piece. In our minds, we naturally envision not just what to paint, but how to paint it, sequencing actions, layering backgrounds, and building structure over time. Understanding and directing this chronological evolution is essential for artists to have real control over their work. The same should hold for methods that aim to recreate the painting process. Yet, current methodologies in painting video synthesis generally suffer from an oversimplification of the underlying spatio-temporal manifold, falling into one of three restrictive paradigms.

Instance Level Optimization: Frameworks like Inverse Painting \cite{chen2024inverse} operate on the assumption that a painting video can be decomposed into an isolated sequence of object-level masks (e.g., sky, mountains). This approach fails in the presence of global artistic techniques, such as multi-instance underpainting, where initial background layers cross object boundaries. Furthermore, its selection mechanism is purely reactive, applying masked inpainting until the mathematical distance to the final image is sufficiently minimized, meaning it possesses no concept of layered composition or user intent.

Progress Based Conditioning: State-of-the-art models like Paints Undo \cite{paintsundo} and Paints Alter \cite{paints_alter} track artwork progression via a singular, global scalar variable. While this allows for coarse backward and forward generation, a 1D scalar cannot capture spatial variance. An artist may completely finish a background element (100\% progress locally) while keeping a foreground subject as a raw sketch (0\% progress locally). Relying entirely on text prompts to fill this spatial steering gap yields highly unpredictable results unless dense reference frames are provided continuously.

Fixed Linear Interpolation: Architectures like Loomis Painter \cite{pobitzer2025loomis} completely omit temporal conditioning inputs. The model implicitly learns a linear progression since the number of frames to generate stayed consistent. This rigid architectural design completely strips away temporal flexibility, making it impossible to adjust the pacing, or order of the generated actions.

In this work we want to address the introduced shortcomings by explicitly grounding the painting video generation task on a fine grained control map that abstracts the painting process as a joint encoding of spatial location and temporal buildup, without considering the stroke color and its physical interaction with the canvas or underlying color, such as color mixing or drying. With the help of such a control mechanism the training data can also be used more efficiently. More importantly the painting process can be viewed in a very detailed manner, coming close to a stroke level generation. Our contributions are as follows: 
\begin{itemize}[noitemsep,topsep=4pt,parsep=0pt,partopsep=0pt]
    \item We introduce the Painting Change Map (PCM), a novel spatio-temporal representation that jointly encodes the spatial location and chronological sequence of a painting process.
    
    \item We propose the Interpolation \& Lineart encoding strategy to seamlessly integrate the PCM into standard RGB video diffusion control videos. At the same time, we show three alternatives, including grayscale interpolation, 24-bit packaging, and historical channel tracking.

    \item We demonstrate that the PCM-based control mechanism can be generated and subsequently used to recreate the creation process of a painting. This two-stage pipeline first allows the model to focus exclusively on the structural painting process, and in the second stage, generates the actual color-based rendering.
    
    \item We provide a comprehensive benchmark comparing our method against generic spatial controls (Canny Edges, Depth) and specialized painting architectures (Paints Undo, Loomis Painter), demonstrating superior localized steering capabilities.
\end{itemize}

%% file: sec/2_related_work.tex
\section{Related work}

\noindent \textbf{Painting process generation.}
Several works have looked at synthesizing the painting process of an input images. Neural painting reconstructs this process through stroke sequences inferred by neural networks~\cite{hu2023stroke, hu2024towards}. Models, like Paint Transformer~\cite{liu2021paint}, use Transformer architectures~\cite{vaswani2017attention,dosovitskiy2020image} to generate stroke parameters. Later methods use diffusion-based neural networks ~\cite{dall2025collaborative, sohl2015deep, ho2020denoising}.

While neural painting models may diverge from real painting processes, pixel-based methods overcome this limitation by synthesizing temporally coherent sequences directly at the pixel level. It has been shown that convolutional approaches~\cite{zhao2020painting} can reconstruct simple workflows, whereas Inverse Painting~\cite{chen2024inverse} iteratively masks the next painting area and inpaints it with a diffusion model.

Recent diffusion-based models improve on these works. ProcessPainter~\cite{song2024processpainter} generates several steps of the painting process as a single image. PaintsUndo~\cite{paintsundo} leverages Stable Diffusion~\cite{rombach2022high} to recreate digital art workflows. Continuing this idea, PaintsAlter~\cite{paints_alter} uses a similar idea to adapt a video diffusion model for the task. Loomis Painter~\cite{pobitzer2025loomis} shows that video diffusion models are well adapted for the task by generating the inverse painting process.

Despite producing realistic workflows, these existing approaches lack mechanisms to actively guide the generation steps. Our work addresses this gap by introducing PCM, an explicit spatio-temporal control mechanism that directly guides where and when the model generates.

\noindent \textbf{Controlling video generation.}
Recent painting process generation models use video generation models like LTX \cite{hacohen2024ltx} or Wan \cite{wan2025wan}. Normally they come in several versions, text-to-video, image-to-video, and video-to-video. PaintsAlter used a video-to-video approach whereas Loomis Painter used an image-to-video approach. For frame level controlled video generation, models normally use a video-to-video approach like in VACE \cite{jiang2025vace} or VideoX-Fun \cite{VideoX_Fun_2026}. For the control maps, several modalities can be used. Some of the most common ones, already found in image generation control mechanisms like ControlNet \cite{zhang2023adding}, include: Canny Edges \cite{canny_edges}, Depth Maps, Normal Maps, and semantic maps. The same mechanisms that have to be shown to work well for image generation can also be used for the video generation models as shown by VideoX-Fun. This enables a pixel level control over the generated frames.

%% file: sec/3_method.tex
\section{Method}

\begin{figure*}[ht]
  \centering
  \includegraphics[width=\linewidth]{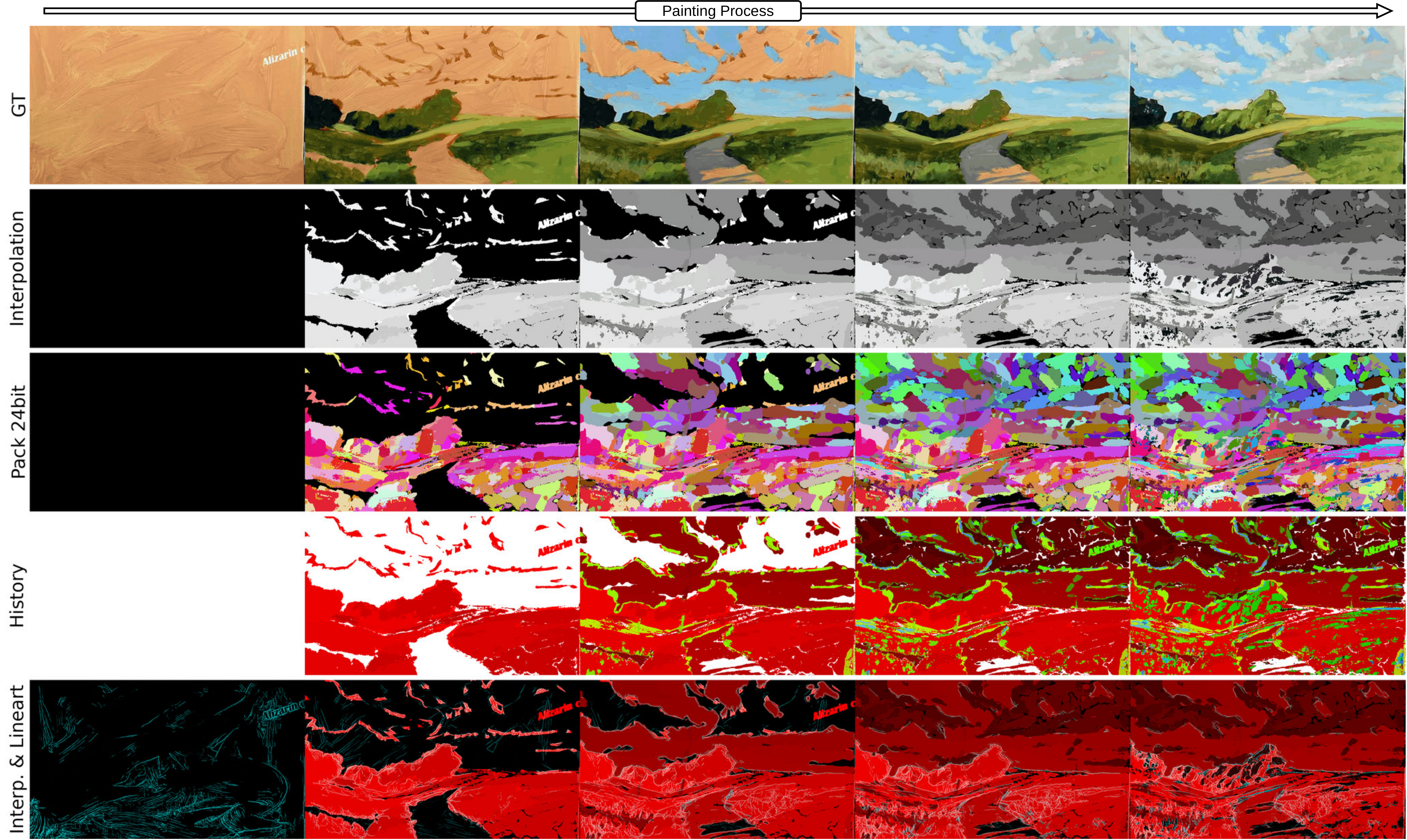}
  \vspace{-6.5mm}
  \caption{\textbf{Comparison of different Painting Change Map modalities.} 
  \textbf{First row:} Ground truth painting process (source video from the ``Painting with Samir'' YouTube channel~\cite{samir_painting_pcm}). 
  \textbf{Second row:} Interpolation-based variant, where earlier painting actions map closer to white and later actions map closer to black. 
  \textbf{Third row:} Pack 24-bit variant, which assigns a distinct color to each individual painting step. 
  \textbf{Fourth row:} History mode, featuring independent interpolation across all three RGB channels. 
  \textbf{Fifth row:} A hybrid approach, fusing the interpolation mechanism (in the R channel) with outlines extracted via a lineart processor (in the GB channels).}
  \label{fig:pcm_overview}
  \vspace{-3mm}
\end{figure*}

The central objective of our framework is to construct a robust control mechanism tailored to the temporal and spatial dynamics of the painting process. Traditional structural guidance methods, such as Canny edges \cite{canny_edges} or Lineart \cite{chan2022lineart}, rely strictly on static feature extraction (e.g., contours and boundaries) to guide generative models. While effective for general image synthesis, these representations do not contain any stroke related information. 

When executing a painting using traditional media (e.g., oil or acrylic), an artist progressively deposits layers of pigment onto a canvas over time. The final output is an accumulation of interrelated variables: the medium's viscosity, structural underpaintings, and individual tool dynamics. Rather than explicitly modeling complex micro-level physical stroke interactions, which we offload to the capacity of our generative backbone, we abstract the painting process as a joint spatiotemporal mapping. This mapping captures exactly where and when geometric changes occur on the canvas.

\subsection{Spatial Painting Change Map}
We formally represent the progression of a painting as a sequence of video frames $\mathcal{V} = \{I_1, I_2, \dots, I_N\}$, where each frame $I_t \in [0, 1]^{C \times H \times W}$. To capture the sequence of artistic actions, we analyze temporal differences across adjacent frames. Our framework accommodates both forward-chronological progression and reverse-chronological order (i.e., the ``un-painting'' sequence starting from a completed artwork down to a blank canvas). We assume an occlusion-free setting where external elements, such as the artist's hands or tools, have been removed.

Given a sequence of consecutive frames, we compute the spatial difference tensor $\Delta V_t = I_{t+1} - I_t$ for $t \in \{1, \dots, N-1\}$. To compress the channel dimension $C$ into a single scalar magnitude map $M_t \in \mathbb{R}^{H \times W}$, we apply a reduction operation. While mean absolute error or maximum absolute deviations are viable, we default to the $L_2$ norm across channels:
\begin{equation}
M_{t,y,x} = \sqrt{\sum_{c=1}^{C} \left( \Delta V_{t,c,y,x} \right)^2}
\end{equation}
A binary change map $B_t \in \{0, 1\}^{H \times W}$ is then instantiated by evaluating $M_t$ against a localized sensitivity threshold $\tau$:
\begin{equation}
B_{t,y,x} = \mathbb{I}(M_{t,y,x} > \tau)
\label{eq:tau}
\end{equation}
where $\mathbb{I}(\cdot)$ denotes the indicator function. The threshold $\tau$ effectively filters out noise and subtle lighting changes.

\subsection{Robustness and Global Motion Filtering}
\label{sec:robustness}
Real-world painting videos introduce physical artifacts that degrade pixel-level delta consistency. Specifically, physical interactions with the canvas cause displacements or camera tremors. This introduces two main types of errors: spatial misalignment over extended periods, and false-positive change detections across the entire frame.

While spatial drift is bounded by the canvas frame boundaries, we mitigate global false-positive triggers by implementing a global motion filter. Given the high temporal resolution of the source video, localized brush strokes occupy only a minor fraction of the total canvas area at any single timestep. Conversely, canvas tremors trigger massive, frame-wide pixel changes. We compute the percentage of altered pixels per frame and nullify frames exceeding a structural safety limit $\theta$:
\begin{equation}
B_t = 
\begin{cases} 
\mathbf{0}_{H \times W} & \text{if } \frac{1}{H \cdot W} \sum_{y=1}^{H}\sum_{x=1}^{W} B_{t,y,x} > \theta \\
B_t & \text{otherwise}
\end{cases}
\label{eq:theta}
\end{equation}
In practice, setting $\theta = 5\%$ successfully purges camera/canvas motion artifacts while preserving large, valid sweeping structural brush strokes. This means that no changes are registered between two frames if more than $5\%$ of the pixels underwent a change.

\subsection{Temporal Encoding and Representation Modes}
To provide a continuous control signal to standard generative pipelines, we construct a cumulative temporal change tensor $\mathcal{L} \in [0, 1]^{T \times H \times W}$. For each spatial coordinate $(y,x)$ at a given timestep $t$, $\mathcal{L}_{t,y,x}$ encodes the normalized index of the most recent timestep $t' \leq t$ where an authentic modification was registered ($B_{t',y,x} = 1$). If no change has occurred up to time $t$, a default baseline fill value is assigned.

To condition downstream diffusion or video generation backbones, we project this multi-channel temporal mapping $\mathcal{L}$ into a standard 3-channel RGB space $\mathcal{C} \in [0, 255]^{T \times H \times W \times 3}$. We introduce four projection modes, as illustrated in \cref{fig:pcm_overview}. While details regarding the Interpolation, Pack 24-bit, and History modes are deferred to Appendix \cref{supp:change_map_modes}, we focus here on a hybrid approach that demonstrated the most robust empirical performance.

\noindent \textbf{Interpolation \& Lineart.}
While pure change maps successfully isolate dynamic events, static structural boundaries remain crucial for stabilizing the control mechanism. Because fine edges are not always detectable from the change maps alone, and given the aforementioned canvas shifts, explicit edge guidance is necessary to reliably anchor the generative model. To address this, we introduce a hybrid approach termed \textit{Interpolation \& Lineart}.

First, we linearly interpolate between two reference boundary colors ($C_{\text{start}}$ and $C_{\text{end}}$) conditioned on the temporal value:
\begin{equation}
\mathcal{C}_{t,y,x} = (1 - \mathcal{L}_{t,y,x}) \cdot C_{\text{start}} + \mathcal{L}_{t,y,x} \cdot C_{\text{end}}
\end{equation}
Setting $C_{\text{start}}$ to black and $C_{\text{end}}$ to white produces a grayscale representation structurally reminiscent of dense depth maps, where brighter values denote more recent artistic interventions. Next, we compute the structural outlines of the targeted frame using a Lineart extractor \cite{chan2022lineart}. Finally, the linear interpolation is fused with the structural data by replacing the green and blue (GB) channels with the extracted lineart, yielding a comprehensive spatiotemporal control signal.

%% file: sec/4_experiments.tex
\begin{table*}[ht]
  \centering
\caption{\textbf{Comparison of Global and Localized Control Performance.} Global metrics are computed across whole unpainting, while localized control averages tasks across instance control categories (e.g., water, tree, meadow). Bold denotes overall best, underline denotes best baseline.}
  \label{tab:unified-method-comparison}
  \resizebox{\textwidth}{!}{%
  \begin{tabular}{l|ccccc|ccccc}
    \toprule
    & \multicolumn{5}{c|}{\textbf{Global Unpainting}} & \multicolumn{5}{c}{\textbf{Local Instance Unpainting}} \\
    \cmidrule{2-6} \cmidrule{7-11}
    \textbf{Method} & \textbf{CLIP↑} & \textbf{DINOv2↑} & \textbf{FID↓} & \textbf{LPIPS↓} & \textbf{MSE↓} & \textbf{CLIP↑} & \textbf{DINOv2↑} & \textbf{FID↓} & \textbf{LPIPS↓} & \textbf{MSE↓} \\
    \midrule
    Depth-anything-v2 \cite{dpeth_anything_v2} & 0.910 & 0.873 & 75.65 & 0.299 & 924.77 & 0.957 & 0.834 & 140.97 & 0.381 & 300.75 \\
    Canny Edges \cite{canny_edges} & \underline{0.923} & 0.908 & 60.23 & 0.226 & 618.11 & \textbf{0.966} & 0.847 & 140.32 & 0.331 & \underline{154.28} \\
    Lineart \cite{chan2022lineart} & 0.922 & \underline{0.917} & \underline{56.92} & \underline{0.213} & \underline{592.81} & 0.965 & \underline{0.867} & \underline{128.30} & \textbf{0.327} & 157.31 \\
    \midrule
    Interpolation & 0.913 & 0.881 & 68.78 & 0.280 & 814.24 & 0.963 & 0.862 & 128.83 & 0.355 & 155.52 \\
    24Bit Packing & 0.910 & 0.875 & 74.06 & 0.296 & 898.68 & 0.963 & 0.860 & 126.37 & 0.358 & 154.92 \\
    History & 0.917 & 0.874 & 68.97 & 0.290 & 819.24 & 0.963 & 0.863 & 126.51 & 0.353 & 152.84 \\
    Interpolation \& Lineart & \textbf{0.926} & \textbf{0.920} & \textbf{55.75} & \textbf{0.207} & \textbf{449.68} & 0.964 & \textbf{0.868} & \textbf{125.91} & \textbf{0.327} & \textbf{149.61} \\
    \bottomrule
  \end{tabular}%
  }
\end{table*}

\section{Experiments}
To evaluate the effectiveness of our proposed framework, we extend the dataset from Loomis Painter~\cite{pobitzer2025loomis} by linearly sampling 1,000 frames from each video. To ensure clean temporal modeling, we replace the original BiRefNet~\cite{BiRefNet} preprocessing pipeline with SAM~3~\cite{carion2025sam3} to segment and remove occlusions (e.g., the artist's hands and tools).

We evaluate our Painting Change Map (PCM) by fine-tuning a Video-to-Video generation model via Low-Rank Adaptation (LoRA)~\cite{hu2022lora}. The model is conditioned on a control video (the change map) and the final completed artwork to synthesize the intermediate painting steps. Specifically, we employ WAN~2.1~Fun~\cite{VideoX_Fun_2026}, an architecture built upon the WAN~2.1~\cite{wan2025wan} video generation foundation model.


Our dataset is the acrylic painting subset consisting of 132 training, 16 test, and 13 validation videos. All results are evaluated on the test split. All variants are trained for 19 epochs on a single NVIDIA A100 (64\,GB) GPU, which equates to approximately 8 hours of training per configuration. We use a learning rate of 1e-4. Following the generation paradigm established by Loomis Painter, the synthesis operates in reverse-chronological order (i.e., ``un-painting'' from the finished motive back to a blank state). Consistent with the underlying base model, our method is trained to generate sequences of 81 frames. To enforce temporal robustness, these sequences are randomly sampled from the 1,000 available video frames during training.

\begin{table}[ht]
  \centering
    \caption{\textbf{Ablation study on filtering thresholds.} We evaluate the impact of the pixel sensitivity threshold $\tau$ (introduced in \cref{eq:tau}) and the global motion filtering limit $\theta$ (introduced in \cref{eq:theta}) using the \textit{Interpolation \& Lineart} modality. Bold numbers indicate the best performance.}
      \label{tab:ablation-pc-threshold}
      \begin{tabular}{cc|cccc}
        \toprule
        \textbf{$\tau$} & \textbf{$\theta$} & \textbf{CLIP↑} & \textbf{DINOv2↑} & \textbf{LPIPS↓} & \textbf{MSE↓} \\
        \midrule
        0.05 & 3.0 & 0.925 & 0.916 & 0.215 & 503.51 \\
        0.05 & 4.0 & 0.925 & 0.916 & 0.214 & 494.96 \\
        \midrule
        0.13 & 3.0 & 0.925 & 0.920 & 0.208 & 458.90 \\
        0.13 & 4.0 & \textbf{0.926} & \textbf{0.920} & 0.207 & 453.25 \\
        0.13 & 5.0 & 0.926 & 0.920 & \textbf{0.207} & 449.68 \\
        \midrule
        0.18 & 3.0 & 0.925 & 0.919 & 0.209 & 456.42 \\
        0.18 & 4.0 & 0.925 & 0.919 & 0.209 & 450.78 \\
        0.18 & 5.0 & 0.924 & 0.920 & 0.208 & \textbf{448.87} \\
        \bottomrule
    \end{tabular}
\end{table}

\begin{figure*}[htbp!]
    \centering

    \begin{subfigure}[b]{\textwidth}
        \centering
        \includegraphics[width=\textwidth]{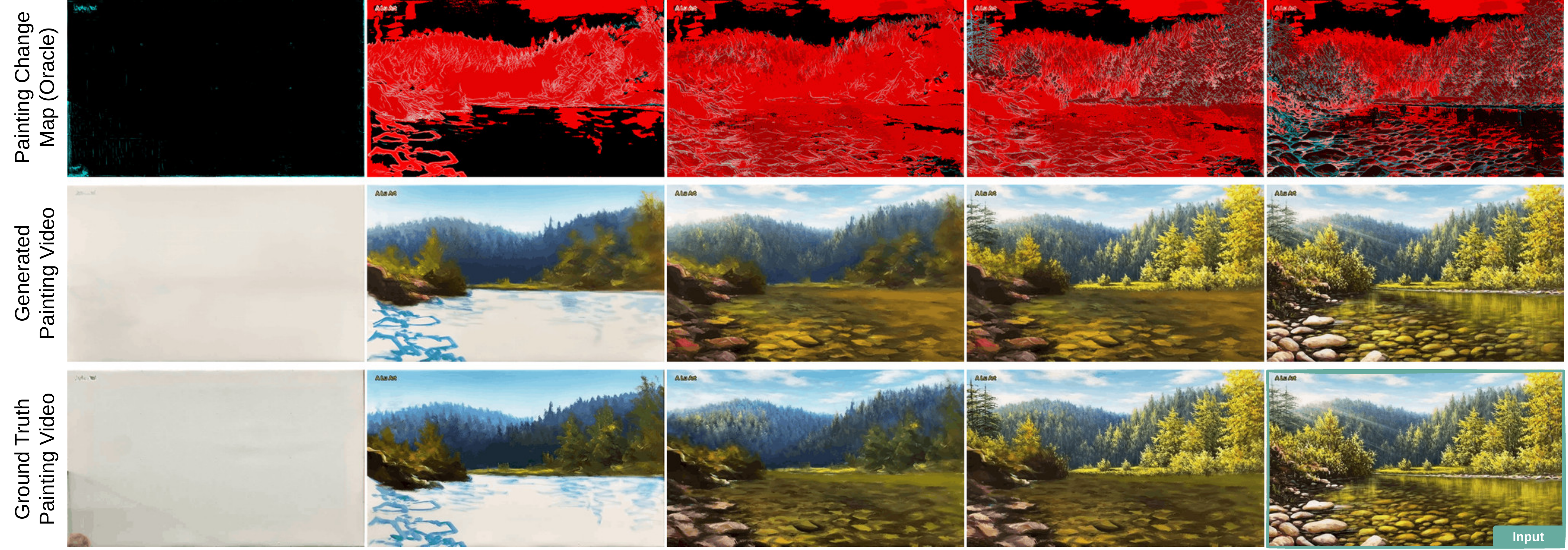}
        \vspace{-4mm}
        \caption{Reconstruction of the painting process using oracle change maps (computed from ground truth video). Source video from the ``A Lu Art'' YouTube channel~\cite{a_lu_art_lake}.}
    \end{subfigure}

    \vspace{3mm}
    \begin{subfigure}[b]{\textwidth}
        \centering
        \includegraphics[width=\textwidth]{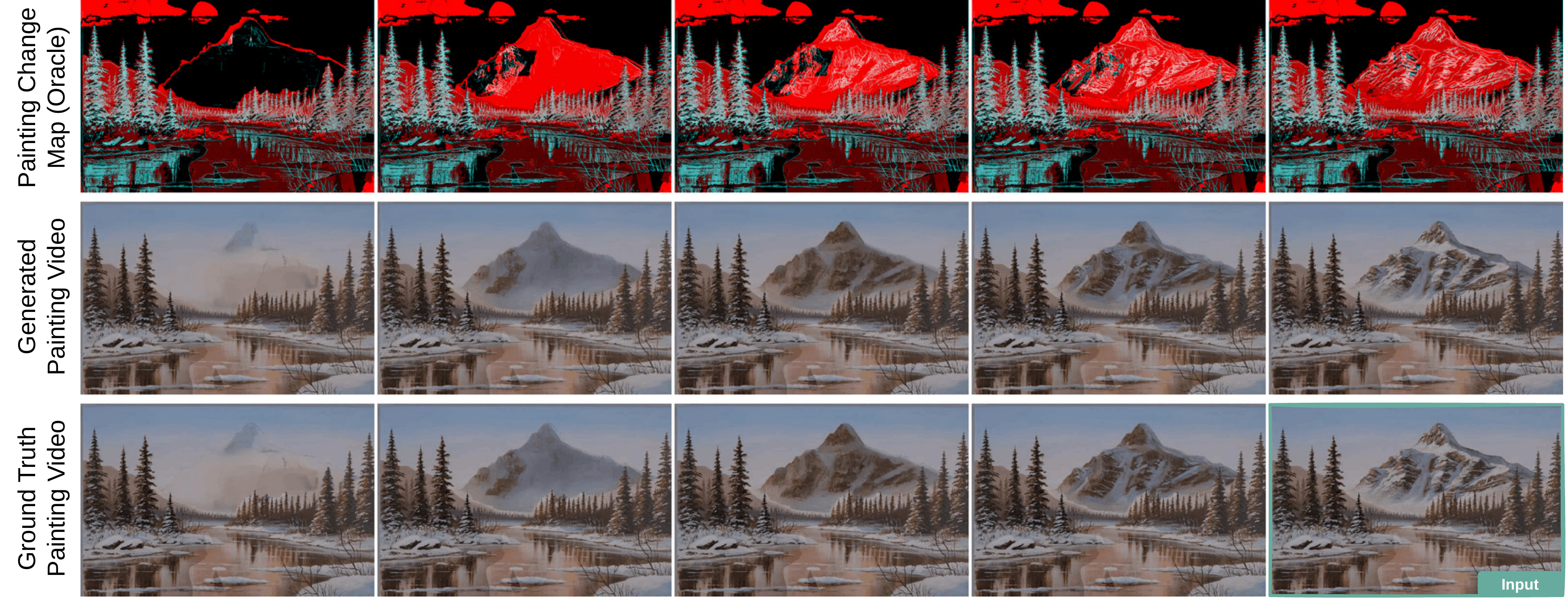}
        \vspace{-4mm}
        \caption{Reconstruction of the painting process focusing only on the mountain and keeping the background untouched. Using oracle change map. Source video from the ``Aham Art'' YouTube channel~\cite{aham_art_snowy_mountain}.}
    \end{subfigure}

    \vspace{-2mm}
    \caption{Visual results of Global and Localized Control Performance. Examples related to \cref{tab:unified-method-comparison}. In (a) the painting process gets reconstructed based on the change map shown in the first row and the input painting in the far right. The change map was computed from the ground truth video shown in the third row. In (b) the setup is the same but only the mountain gets unpainted keeping the background stable. This shows the generalization capabilities of the control mechanism.}
    \label{fig:showcase}
\end{figure*}

\begin{figure*}[htbp!]
    \centering

    \vspace{3mm}
    \begin{subfigure}[b]{\textwidth}
        \centering
        \includegraphics[width=\textwidth]{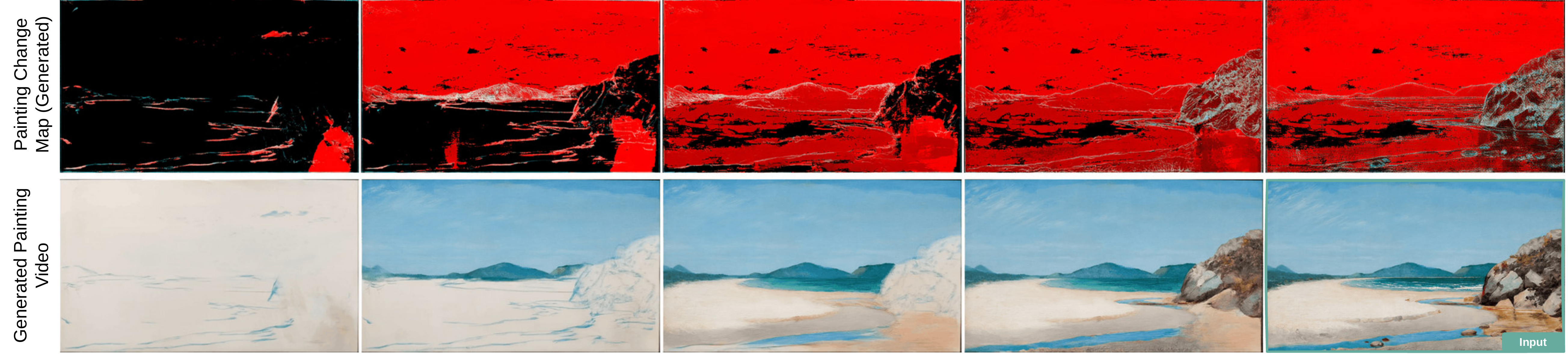}
        \vspace{-4mm}
        \caption{Reconstruction of the Painting process. \textit{Seascape, Guarujá} by Almeida Júnior, Public Domain, via Wikimedia Commons.}
    \end{subfigure}

    \vspace{3mm}
    \begin{subfigure}[b]{\textwidth}
        \centering
        \includegraphics[width=\textwidth]{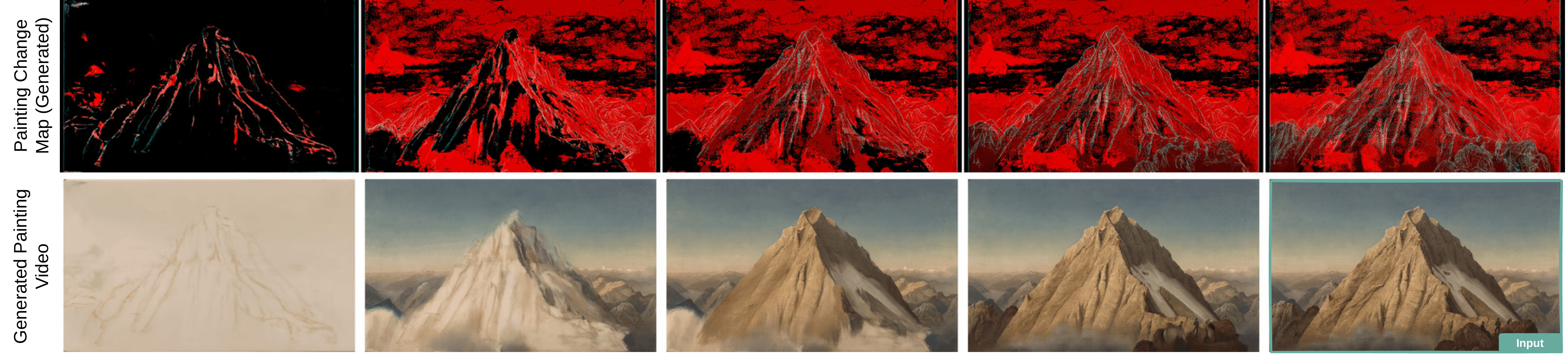}
        \vspace{-4mm}
        \caption{Reconstruction of the Painting process. \textit{Peak of Mount Triglav} by Marko Pernhart, Public Domain, via Wikimedia Commons.}
    \end{subfigure}

    \vspace{-2mm}
    \caption{Reconstruction of the Painting process using in the wild paintings. The input is the right most image. First the change map gets generated based on the input. In a second step the painting video is generated based on the input image and the generated change map.}
    \label{fig:showcase_in_the_wild}
\end{figure*}

\subsection{Baselines}
We evaluate against three prominent structural control baselines, each fine-tuned under identical training configurations as our methods:

\noindent \textbf{Depth-Anything-v2} \cite{dpeth_anything_v2}: A robust depth estimation model. The generation pipeline is conditioned on the estimated depth map corresponding to each target frame.

\noindent \textbf{Canny Edges} \cite{canny_edges}: Standard high-frequency edge maps extracted from the target frames to guide spatial boundaries.

\noindent \textbf{Lineart} \cite{chan2022lineart}: Artistic outlines optimized specifically for line drawings and painterly illustrations.

\subsection{Reconstructing the painting process}
First, we show the general capabilities of recreating the global painting process. For a rigorous upper-bound comparison, all experiments utilize ground-truth (oracle) control maps during inference.

Quantitative results are reported in \cref{tab:unified-method-comparison} (Global Unpainting). Crucially, we observe that standalone temporal change maps (\textit{Interpolate}, \textit{24-Bit Packing}, and \textit{History}) underperform relative to pure edge-based baselines like \textit{Canny} and \textit{Lineart}. We attribute this behavior to two primary factors. 

First, because our generation paradigm conditions the network on the final completed painting, outlines provide a strong structural prior for cross-frame alignment when reversing the painting sequence. Second, as detailed in \cref{sec:robustness}, movements of the physical camera or canvas introduce spatial drift over time. Ground-truth structural baselines implicitly correct for these shifts by offering fixed geometric anchors in every frame. In contrast, pure temporal change maps isolate localized stroke updates; without explicit static structures, the model struggles to infer global motion from isolated stroke trajectories alone.

By explicitly addressing this limitation, our combined approach Interpolation \& Lineart bridges both paradigms. By leveraging the stroke progression and the stability of static Lineart, this variant registers substantial performance gains, outperforming all baselines and establishing a new state-of-the-art across every evaluated metric (notably reducing MSE from 592.81 to 449.68).

\subsection{Local Instance Control}

To evaluate the versatility beyond global unpainting, we investigate our framework's capacity for fine-grained localized control. This represents a zero-shot scenario, as it diverges from our training strategy. Specifically, we isolate distinct semantic instances (e.g., trees, lakes, mountains) and task the model with unpainting only the targeted region while preserving the static background. Our evaluation spans seven distinct instance categories, conditionally sampled based on their presence within each source painting.

The quantitative results for this task are reported in \cref{tab:unified-method-comparison}. To ensure a rigorous localized evaluation, metric computation is strictly confined to the masked instance areas. For DINOv2 and LPIPS, masks are applied directly during feature extraction, whereas for CLIP, FID, and MSE, non-target regions of the frames are masked with a white background prior to evaluation. Our proposed \textit{Interpolation \& Lineart} modality maintains a competitive edge. It achieves the best overall performance in FID, MSE, and DINOv2, while tying the Lineart baseline in LPIPS.


\subsection{Qualitative Results}
We show qualitative results for the global painting process in \cref{fig:showcase} (a) where the painting change map was generated based on the ground truth video. The same holds for  \cref{fig:showcase} (b) where the painting process only touches the mountain. This represents a zero-shot task since during training only the global process was seen.

In \cref{fig:showcase_in_the_wild}, we show results for in-the-wild-paintings. For these we do not have ground truth painting videos and the painting change map was generated by our model. Based on the generated change maps and the reference painting the actual painting process gets reconstructed.


\subsection{Ablation of Spatiotemporal Thresholds}
The fidelity of our PCM relies on two hyperparameters: the localized spatial sensitivity threshold $\tau$, which separates intentional brushstrokes from sensor noise and minor illumination shifts, and the global motion filtering threshold $\theta$, which isolates camera tremors and canvas displacements. To assess the downstream impact of these hyperparameter configurations on generative performance, we conduct a grid-search ablation study using the \textit{Interpolation \& Lineart} modality as our baseline. Quantitative results across semantic and perceptual metrics are detailed in \cref{tab:ablation-pc-threshold}.

Our analysis reveals that performance is highly robust across a reasonable operational window, yet sensitive at extreme values. Setting a low sensitivity threshold ($\tau = 0.05$) forces the model to register ambient noise and compression artifacts as valid changes. This introduces spurious artifacts into the cumulative temporal tensor, leading to degraded perceptual metrics (e.g., LPIPS increases to $0.214$ and MSE rises sharply to $494.96$). Conversely, over-filtering with a high threshold ($\tau = 0.18$) aggressively strips away genuine, fine-grained stroke actions, causing a subtle drop in semantic metrics like CLIP and DINOv2 as the downstream model loses temporal context. The configuration of $\tau = 0.13$ consistently emerges as the optimal point.

Regarding the global motion filter, variations in $\theta$ demonstrate a trade-off between structural stability and motion purging. A highly restrictive threshold ($\theta \leq 3.0\%$) prematurely discards frames containing large, authentic, sweeping brush strokes, which marginally hurts reconstruction metrics. Ultimately, the pairing of $\tau = 0.13$ and $\theta \in [4.0, 5.0]\%$ yields the best equilibrium.

\begin{table*}[htpb!]
  \centering
  \caption{\textbf{Quantitative Comparison with State-of-the-Art Painting Generation Methods.} Metrics evaluate temporal reconstruction fidelity, structural similarity, and computational efficiency. $\dagger$ denotes our method using ground-truth (oracle) change maps to demonstrate the performance upper bound. Bold numbers indicate the best non-oracle performance, underline the second best. Params, Latency, and FPS are only related to the generative part of each method.}
  \label{tab:art_evaluation}
  \begin{tabular}{l|ccccc|cccc}
    \toprule
    \textbf{Method} & \textbf{CLIP $\uparrow$} & \textbf{DINOv2 $\uparrow$} & \textbf{FID $\downarrow$} & \textbf{LPIPS $\downarrow$} & \textbf{MSE $\downarrow$} & \textbf{Params} & \textbf{Latency (s)} & \textbf{FPS} \\
    \midrule
    Inverse Painting ~\cite{chen2024inverse} & 0.69 & 0.27 & 287.58 & 0.77 & 10007.06 & 900M & 333.8 & 0.16 \\
    ProcessPainter~\cite{song2024processpainter} & 0.77 & 0.51 & 223.16 & 0.60 & 5253.44 & 1.3B & 1.7 & 4.71  \\
    PaintsUndo~\cite{paintsundo} & 0.76 & 0.50 & 209.22 & 0.66 & 9448.22 & 900M & 129.6 & 0.62 \\
    Loomis Painter~\cite{pobitzer2025loomis} & \textbf{0.88} & \textbf{0.81} & \textbf{101.72} & \textbf{0.44} & \underline{3152.96} & 14B & 428.3 & 0.19 \\
    \midrule
    Interpolation \& Lineart & \underline{0.87} & \underline{0.74} & \underline{119.53} & \underline{0.45} & \textbf{3127.13} & 5B + 1.3B & 130.74 & 0.62 \\
    \rowcolor{gray!20}
    Interpolation \& Lineart$^\dagger$ & 0.93 & 0.92 & 55.75 & 0.21 & 449.68 & 1.3B & 96.4 & 0.84 \\
    \bottomrule
  \end{tabular}
\end{table*}

\subsection{Comparison to Painting Generation Methods}
To demonstrate the broader utility and generalizability of our control mechanism, we evaluate our framework against state of the art approaches for painting process reconstruction. Unlike the oracle evaluations in previous sections, recreating target painting requires generating its corresponding control signals from scratch.

To achieve this, we propose a decoupled, two-stage inference pipeline. In the first stage, we fine-tune a Wan2.1 Image-to-Video (I2V) 5B model \cite{wan2025wan} to predict the change maps. Given a painting, this model synthesizes an 81-frame sequence of corresponding \textit{Interpolation \& Lineart} change maps. In the second stage, these generated maps serve as the conditioning signal for our previously trained 1.3B Video-to-Video generation model to synthesize the final unpainting sequence.

We compare our pipeline against four prominent baselines: Inverse Painting \cite{chen2024inverse}, ProcessPainter \cite{song2024processpainter}, PaintsUndo \cite{paintsundo}, and Loomis Painter \cite{pobitzer2025loomis}. Quantitative results are detailed in \cref{tab:art_evaluation}. To establish a rigorous performance ceiling, we also report our method using ground-truth change maps (denoted by $\dagger$).

Our full generative pipeline generalizes well to unseen artwork, nearly tying the leading baseline, Loomis Painter, in semantic consistency (0.87 vs. 0.88 CLIP) and perceptual fidelity (0.45 vs. 0.44 LPIPS), while strictly outperforming it in structural pixel accuracy (3127.13 vs. 3152.96 MSE). Crucially, while Loomis Painter deterministically samples the same 81 uniformly spaced frames during training, our randomized sampling strategy spans all 1,000 available video frames. This prevents dataset underutilization and inherently improves temporal robustness. Furthermore, our oracle variant ($\dagger$) outperforms all existing baselines by a substantial margin across every metric, highlighting an exceptionally high performance upper bound as predictive backbones mature.

Beyond synthesis quality, our decoupled framework offers a significant latency advantage over heavy architectures. By more than tripling the throughput to 0.62 FPS compared to Loomis Painter's 14B model, our approach proves highly efficient. Because our components execute sequentially rather than concurrently, the memory footprints do not stack. The peak hardware requirement is strictly isolated to 25.84 GB of VRAM during the first stage, dropping to 17.04 GB of VRAM during the second stage. This low memory overhead makes our framework highly accessible for standard mid-tier hardware configurations and renders the pipeline trivially parallelizable. All experiments were conducted on a single NVIDIA A100 SXM4 64GB.

%% file: sec/5_conclusion.tex
\section{Conclusion}
\label{sec:conclusion}

In this work, we presented a novel framework for controllable painting video synthesis that decouples abstract painting actions from the complex physical properties of traditional media. By shifting the generative focus away from modeling intricate interactions like color mixing, paint viscosity, and canvas textures, our framework isolates the core spatiotemporal progression of individual brushstrokes. Through the introduction of the Painting Change Map (PCM) and a decoupled, two-stage inference pipeline, our approach achieves strong evaluation results across both global unpainting benchmarks and localized instance-level control tasks, establishing a high performance upper bound.

A central finding of our study is the critical importance of structural outlines when guiding the generation process. For video material captured from real world painting processes, minor canvas displacements and camera tremors are structurally inevitable. While isolated strokes capture localized actions, they struggle to infer global trajectories amidst these shifts. Fusing temporal progression with explicit lineart boundaries acts as a vital geometric anchor, enabling the diffusion model to maintain cross-frame alignment.

\noindent\textbf{Limitations and Future Work.}
The primary limitations of our approach stem directly from the characteristics of the source dataset. Despite implementing global motion filtering, physical canvas movements mean that strokes painted in the past can gradually lose synchronization and desynchronize from the current frame geometry over extended periods. While this issue would be entirely absent in ideal, perfectly stabilized data, real-world conditions necessitate explicit tracking. Consequently, an promising direction for future work is to integrate automated canvas alignment or keypoint rectification modules into the data pipeline to proactively eliminate spatial drift.

Additionally, we aim to further analyze how much information is carried by the change maps alone. Can the art medium, style or even the painter itself be identified through it? Currently, this direction is constrained by the scarcity of sufficiently diverse and large-scale datasets.

\section{Acknowledgement}
This work was supported by the EU Horizon project “ELIAS - European Lighthouse of AI for Sustainability” (No. 101120237) and the FIS project GUIDANCE (Debugging Computer Vision Models via Controlled Cross-modal Generation) (No. FIS2023-03251).

We acknowledge ISCRA for awarding this project access to the LEONARDO \cite{leonardo} supercomputer, owned by the EuroHPC Joint Undertaking, hosted by CINECA (Italy) .

%% file: sec/6_supplementary.tex
\clearpage
\section{Supplementary}

\subsection{Representation Modes Change Map}
\label{supp:change_map_modes}

\paragraph{Interpolation.} 
The most direct transformation applies a linear interpolation between two reference boundary colors ($C_{\text{start}}$ and $C_{\text{end}}$) based on the temporal value:
\begin{equation}
\mathcal{C}_{t,y,x} = (1 - \mathcal{L}_{t,y,x}) \cdot C_{\text{start}} + \mathcal{L}_{t,y,x} \cdot C_{\text{end}}
\end{equation}
Setting $C_{\text{start}}$ to black and $C_{\text{end}}$ to white produces a grayscale representation structurally reminiscent of dense depth maps, where brighter values denote more recent artistic interventions.

\paragraph{24-bit Packing.} 
To circumvent scalar quantization constraints and map deep temporal histories without losing fine-grained fidelity, we distribute the scalar values losslessly across the full 24-bit RGB space. The normalized continuous coordinate is mapped to an integer domain $V = \lfloor \mathcal{L}_{t,y,x} \times (2^{24}-1) \rfloor$, which is bitwise partitioned into distinct 8-bit color channels:
\begin{equation}
    \begin{split}
        R = (V \gg 16) \,\&\, 255 \\
        G = (V \gg 8) \,\&\,  255 \\
        B = V \,\&\, 255
    \end{split}
\end{equation}
This approach supports up to $16.7$ million steps. While this provides exceptionally high temporal precision, it introduces complex chromatic discontinuities that can complicate pattern recognition.

\paragraph{History.}
This configuration models local temporal trajectory by tracking a sliding history window of updates. Rather than mapping a single event, the $R$, $G$, and $B$ channels are mapped to the standard linear interpolations of the three most recent sequential modifications at that specific pixel location. This exposes canvas evolution directly within a single multi-channel tensor.

\paragraph{Interpolation \& Lineart.}
Pure change maps highlight dynamic events rather than static edge boundaries. To integrate both paradigms, we compute structural outlines of the targeted frame using a Lineart extractor \cite{chan2022lineart}. As the base map we use the Interpolation mode introduced above and replace the last two channels (GB) with the edge maps extracted by the Lineart model.